\documentclass{llncs}
\usepackage{subfigure}
\usepackage{graphicx}
\usepackage{amsmath}

\begin{document}

\title{\bf Parametric Curve Segment Extraction by Support Regions}

\author{Cem \"{U}nsalan}
\institute{Department of Electrical and Electronics Engineering,\\
Yeditepe University, \.{I}stanbul 34755, Turkey.\\
unsalan@yeditepe.edu.tr\\}

\maketitle

\begin{abstract}
We introduce a method to extract curve segments in parametric form from the image directly using the Laplacian of Gaussian (LoG)
filter response. Our segmentation gives convex and concave curves. To do so, we form curve support regions by grouping pixels of
the thresholded filter response. Then, we model each support region boundary by Fourier series and extract the corresponding
parametric curve segment.
\end{abstract}

\section{Introduction}

High level computer vision tasks, such as object recognition and scene understanding requires primitives to be extracted from the
image. One such primitive is the edge information. State of the art edge detectors proposed in the literature can be found in a
standard image processing book such as \cite{Boyle1}. One of the best known edge detectors is Canny's method \cite{Canny1}.

As the edge pixels are detected, they can be represented as curves. For high level processing, these curves need to be segmented
and represented in a standardized form. The literature is vast on curve segmentation
\cite{Boyer2,Rosin2,Lindeberg1,Rosin6,Cronin1}. Most of these methods require the candidate curve segment pixels to be extracted
first, either by edge detection or a similar preprocessing. Then, these pixels are grouped by a given criteria to form the
corresponding curve segments.

These curve segments should also be represented in parametric form. One way of parametric representation is polygonal
approximation. Some of the existing work on this problem is given in references \cite{Rosin7,Davis1,Bao1}. Most of these methods
require curve segments to be extracted first by a preprocessing step. Some of these methods are also based on the area
information, such that the contours in the binary image is represented in terms of polygons. However, these methods require a
thresholding step to be applied to grayscale images.

In this study, we extract the trigonometric polynomial representation of curve segments in the image by a method combining edge
pixel grouping and curve segmentation. Our method is similar to straight line extraction using line support regions introduced by
Burns~\emph{et. al.}~\cite{Riseman1}. They grouped pixels of contiguous gradient directions and formed support regions. Then they
obtained straight line segments from these. This method also combines steps in parametric curve segment extraction. However, the
final segments extracted are only straight lines and the method is not suitable for a general curve segment extraction. Another
drawback of their method is its computational cost because of the pixel grouping step.

To overcome these problems, we extend the idea to represent the general parametric curve segments by dividing them into convex
and concave sections. Hence, our method can handle curved and straight line segments in the image. We group pixels to form curve
support regions by the Laplacian of Gaussian (LoG) filter response (by extending its usage beyond zero crossing edge detection).
Then, we obtain the corresponding parametric curve segment by fitting a Fourier series to its boundary. Next, we explore our
curve segmentation and parametrization method step by step, starting with mathematical derivation followed by experimental
results.

\section{Curve Support Region (CSR) Extraction}

Zeros of the second derivative of a one dimensional function indicate its inflection points. Therefore, zeros of the second derivative can
be used to distinguish convex and concave parts of the function. Haralick and Shapiro \cite{Shapiro1} point out that ``The isotropic
generalization of the second order derivative to two dimensions is the Laplacian''. Based on Haralick and Shapiro's definition, the
Laplacian should have the same property (partitioning the convex and concave parts) for two dimensional functions (such as the brightness
function of an image). In this study, we give examples showing that this is indeed the case. Therefore, this property will lead to direct
curve segmentation, partitioned into convex and concave parts, from the image at hand.

\subsection{An Example in Continuous Domain}

We next give an example in continuous domain. We model an intensity change (in an image brightness function) having a sinusoidal
shape, consisting of convex and concave segments. This example gives insight into the mathematical background of the proposed
method. As we are working in continuous domain, we obtain curve segments mathematically.

Let a curve on a continuous image brightness function be:

\begin{equation} \label{eq:curvex1}
y -\sin(x)+\rho=0
\end{equation}

\noindent where $\rho$ is a constant. To generalize the curve in 2D, let's add a $\theta$ radian rotation.

\begin{equation} \label{eq:rotatedcurve}
\begin{bmatrix}
y_r\\
x_r\\
\end{bmatrix}
=
\begin{bmatrix}
\cos(\theta)&-\sin(\theta)\\
\sin(\theta)&\cos(\theta)\\
\end{bmatrix}
\times
\begin{bmatrix}
y\\
x\\
\end{bmatrix}
=
\begin{bmatrix}
\alpha&-\beta\\
\beta&\alpha\\
\end{bmatrix}
\times
\begin{bmatrix}
y\\
x\\
\end{bmatrix}
\end{equation}

\noindent where $\alpha=\cos(\theta)$ and $\beta=\sin(\theta)$. Our rotated curve becomes

\begin{equation} \label{eq:curvex2}
y_r -\sin(x_r)+\rho=0
\end{equation}

Now, let's define an image brightness function $B(x,y)$ as

\begin{equation} \label{eq:bright1}
B(x,y)=b_1+\Delta b u(y_r-\sin(x_r)+\rho)
\end{equation}

\noindent where $b_1$ is a brightness value, $\Delta b$ is the contrast, and $u(\cdot)$ is the unit step function. The brightness function
becomes

\begin{equation} \label{eq:bright2}
B(x,y)=b_1+\Delta b u(\alpha y -\beta x  -\sin(\beta y + \alpha x )  +\rho)
\end{equation}

The Laplacian over $B(x,y)$ is calculated as

\begin{equation} \label{eq:laplacian1}
\nabla^2 B= \frac {\partial^2 B} {\partial x^2} +\frac {\partial^2 B} {\partial y^2}
\end{equation}

\noindent  where the second order partial derivatives are

\begin{equation} \label{eq:derivative2}
\begin{split}
\frac {\partial^2 B} {\partial x^2}=&\Delta b [\alpha^2 \sin(\beta y+ \alpha x) \delta (\alpha y -\beta x  -\sin(\beta y + \alpha x) +\rho)\\
&+(-\beta -\alpha \cos(\beta y + \alpha x))^2\delta'(\alpha y -\beta x -\sin(\beta y+\alpha x) +\rho)]
\end{split}
\end{equation}

\begin{equation} \label{eq:derivative4}
\begin{split}
\frac {\partial^2 B} {\partial y^2}=& \Delta b [\beta^2 \sin(\beta y+\alpha x) \delta (\alpha y -\beta x  -\sin(\beta y +\alpha x )+\rho)\\
&+(\alpha -\beta \cos(\beta y + \alpha x ))^2\delta'(\alpha y -\beta x  -\sin(\beta y +\alpha x +\rho)]
\end{split}
\end{equation}

\noindent We obtain

\begin{equation} \label{eq:laplacian2}
\begin{split}
\nabla^2 B=&\Delta b [\sin(\beta y +\alpha x ) \delta (\alpha y -\beta x -\sin(\beta y + \alpha x )+\rho)\\
&+(1+\cos^2(\beta y+ \alpha x ))\delta'(\alpha y -\beta x  -\sin(\beta y +\alpha x )+\rho)]
\end{split}
\end{equation}

Here, we take $\delta(t)$ to be nonzero only at $t=0$ and $\delta'(t)$ to be nonzero only as $t\rightarrow0^+$ (with negative response) and
$t\rightarrow0^-$ (with positive response). This is consistent with the zero crossing edge detection in computer vision, where we have a
zero response on the edge and nonzero responses (with different polarities) on the two sides of the edge.

If we want to localize edges in this brightness function, we should equate

\begin{equation} \label{eq:delta1}
\delta (\alpha y -\beta x  -\sin(\beta y + \alpha x ) +\rho)=1
\end{equation}

\noindent This is satisfied only for

\begin{equation} \label{eq:delta2}
\alpha y -\beta x -\sin(\beta y+\alpha x) +\rho=0
\end{equation}

\noindent which is the expanded form of the rotated curve in Eqn.~\ref{eq:curvex2}. Therefore, the $\delta(\cdot)$ response in
the Laplacian of $B(x,y)$ gives the edge locations mathematically.

The magnitude of $\delta(\cdot)$ on the curve is $\sin(x_r)$. This function is zero only for $x_r=n\pi$, where $n$ is any integer
number. On the curve, these $x_r$ values lead to $y_r=-\rho$. Therefore, points $(n\pi,-\rho)$ correspond to zeros of the
Laplacian of the brightness function in rotated coordinates. These points are the actual inflection points of the rotated curve.
Hence, finding zeros of the Laplacian actually corresponds to finding the inflection points of the 2D function.

More important to us is the response of $\delta'(\cdot)$ around zero. Let's consider the positive response obtained from $\delta'(0^-)$. In
order to obtain the nonzero response for this function, we must have

\begin{equation} \label{eq:doublet1}
\alpha y -\beta x -\sin(\beta y+\alpha x) +\rho=\lim_{\epsilon \rightarrow 0^-} \epsilon
\end{equation}

\noindent This equation indicates that the curve can be located from the $\delta'(0^-)$ response in $\epsilon$ distance from the actual
location as in Eqn.~\ref{eq:delta2}.

The magnitude of the positive response of $\delta'(\cdot)$ is $1+cos^2(x_r)$. Thresholding this function around two divides the response
into subparts separated by points $(n\pi,-\rho)$. These are the inflection points on the rotated curve. Therefore, these subparts
correspond to separate convex and concave curve segments. Similar to this example, obtaining the nonzero $\delta'(\cdot)$ response and
thresholding it will give the segmented curves in continuous domain functions.

\begin{figure*}[htbp]
\centering

\subfigure[$B(x,y)$]{\includegraphics[height=1.2 in]{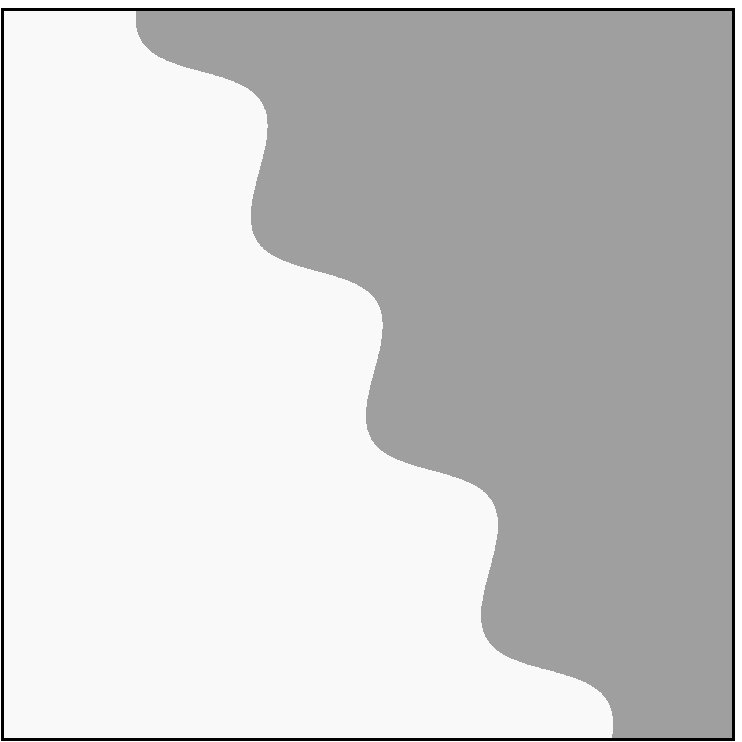}} \hspace{.2 in}\subfigure[Curve
segments]{\includegraphics[height=1.2 in]{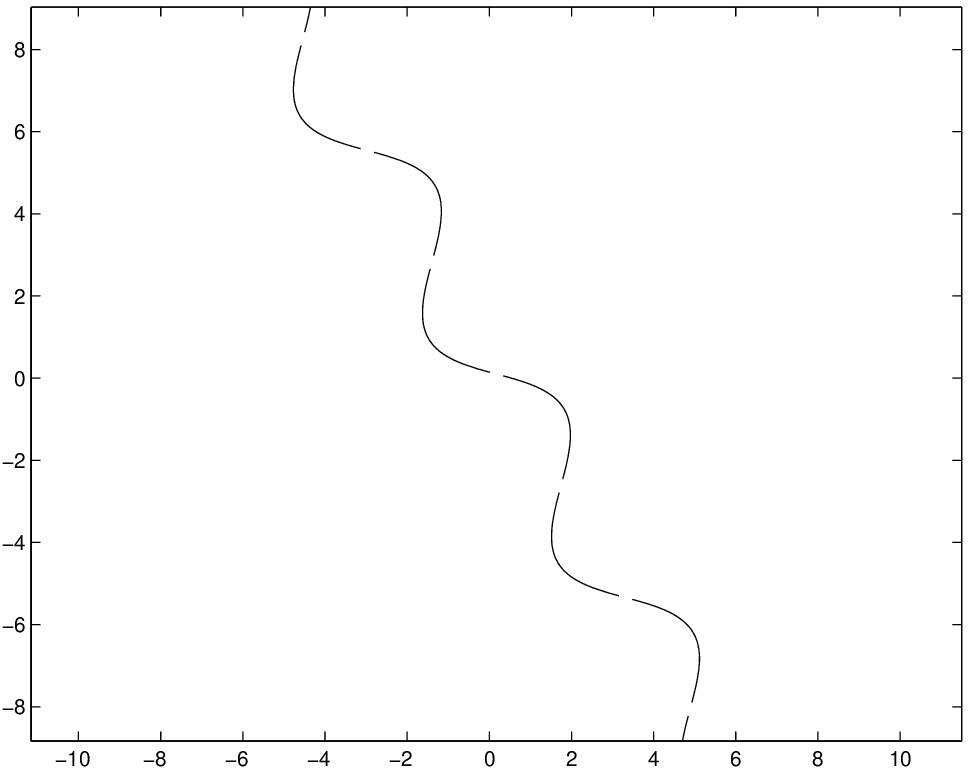}} \caption{A curve segmentation example in continuous
domain}\label{fig:curve}
\end{figure*}

We simulate the brightness function, $B(x,y)$ in Fig.~\ref{fig:curve}~(a) with a $\pi/3$ radians rotation and $\rho=0.2$. The
curve segments are given in Fig.~\ref{fig:curve}~(b). As can be seen, the Laplacian performs the same operation in two dimensions
as does the second derivative in one dimension. Therefore, we can extract curve segments directly by using the $\delta'(\cdot)$
part of the Laplacian of a continuous domain function.

\subsection{An Example in Discrete Domain}

To apply the above idea on digital images, we need to obtain the discrete version of the Laplacian operator with a possible
smoothing applied on the image. Marr and Hildreth~\cite{Hildreth1} introduced the rotation invariant Laplacian of Gaussian (LoG)
filter for a similar purpose. They used the Laplacian of a smoothed image to extract edges in a zero crossing sense. In this
method, the Laplacian operator performs the partial second derivative calculations and the Gaussian filter performs smoothing on
the image. By the property of convolution, both smoothing and Laplacian operations can be represented as one filtering operation,
hence the LoG filter:

\begin{equation} \label{eq:logfilter}
f(x,y) = \frac {1} {\pi\sigma^4} \left[\frac {x^2+y^2} {2\sigma^2} -1 \right] \exp(-\frac {x^2+y^2} {2\sigma^2})
\end{equation}

\noindent where $\sigma$ is the scale parameter (for smoothing) of the LoG filter. The scale parameter leads to a natural scale space
property for curve extraction and segmentation.

\begin{figure*}[htbp]
\centering \subfigure[Rectangle]{\includegraphics[width=1.1 in]{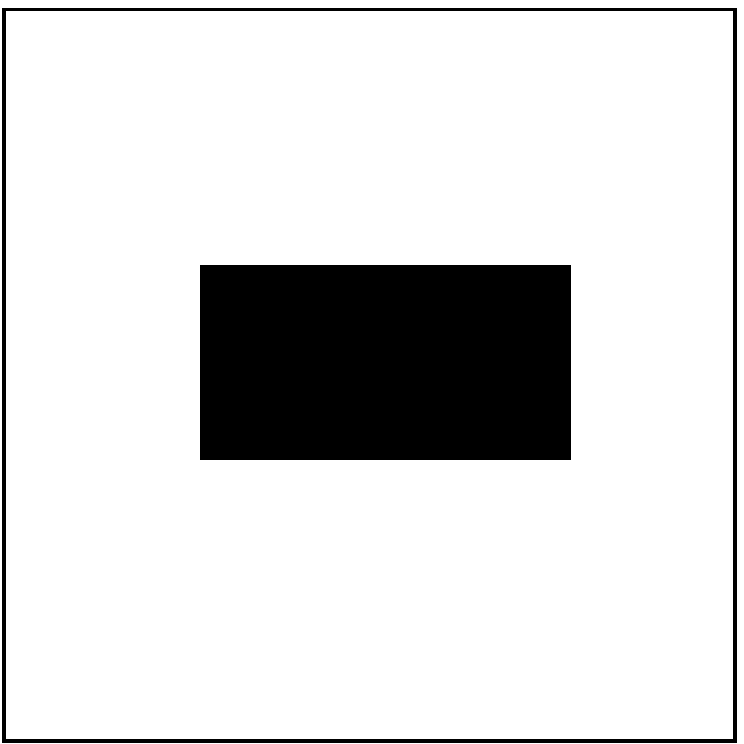}} \subfigure[LoG
response]{\includegraphics[width=1.1 in]{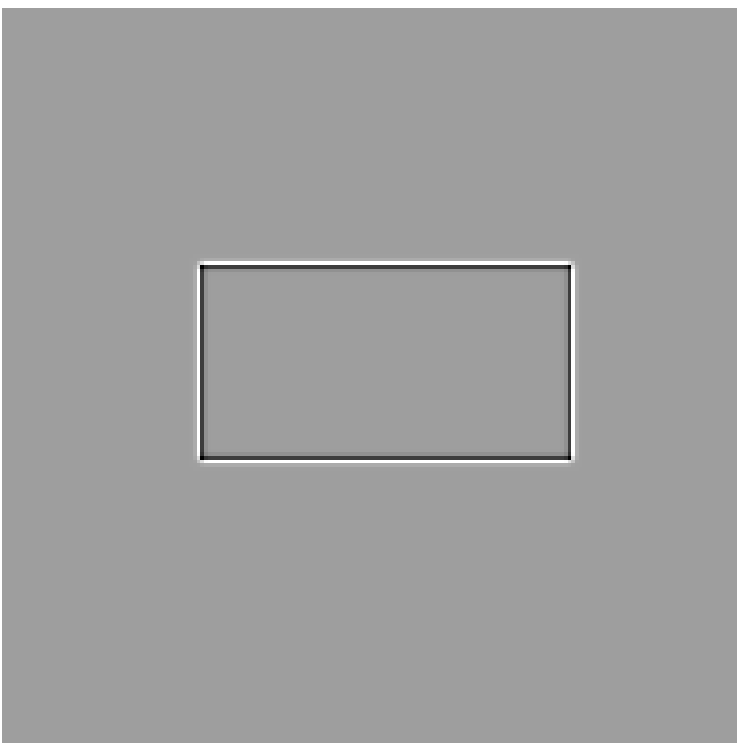}} \subfigure[CSR extracted]{\includegraphics[width=1.1
in]{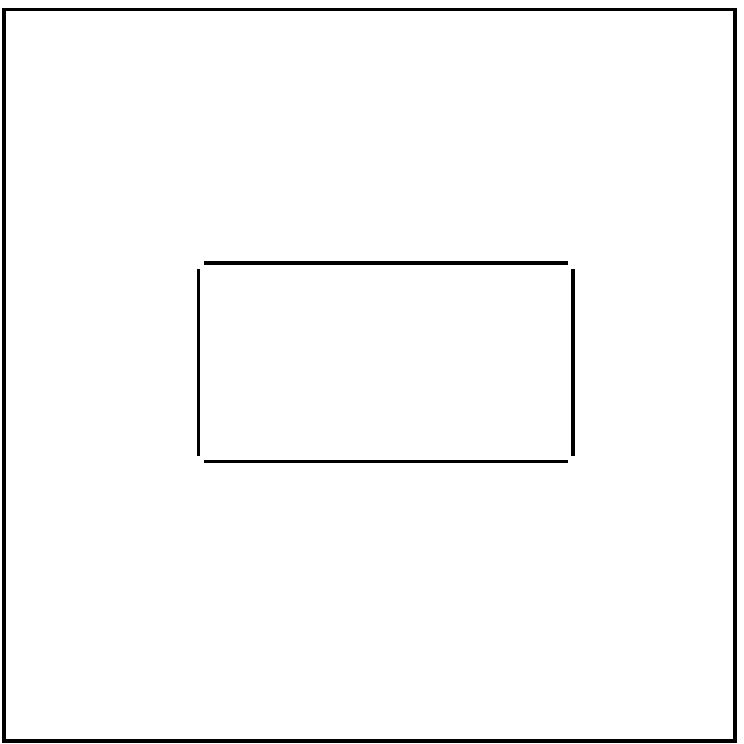}} \subfigure[Curve segments]{\includegraphics[width=1.1 in]{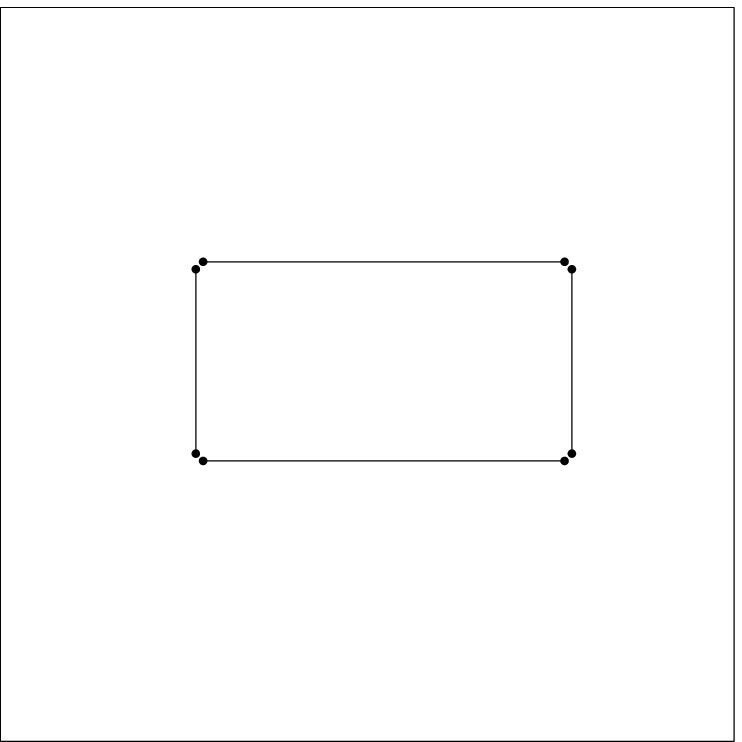}}

\caption{Curve support regions and curve segments over the rectangle image} \label{fig:rectangle}
\end{figure*}

We next apply the LoG filter to obtain the positive and negative responses similar to the  $\delta'(\cdot)$ response above. We
label pixels having responses higher than a threshold. Connected components analysis on these pixels leads to curve support
regions of the image. We give an example of this procedure on an image containing a rectangle shape in
Fig.~\ref{fig:rectangle}~(a).

In Fig.~\ref{fig:rectangle}~(b) the LoG filter response has three levels as: the zero level (gray), high level (bright), and the
low level (black). To form curve support regions, we threshold this response and take only the high valued pixels (bright pixels
in Fig.~\ref{fig:rectangle}~(b)). Curve support regions for the rectangle image are given in Fig.~\ref{fig:rectangle}~(c). As can
be seen, curve support regions indicate four separate curve segments for this shape. We will consider obtaining the final
parametric representation of the curve segments from these support regions next.

\section{Parametric Curve Segment Extraction}

To extract the final parametric curve segment from the corresponding support region, we represent the outer boundary of each
support region using its Fourier descriptors. In a previous study, we applied a similar strategy to extract straight line
segments from the line support regions \cite{Unsalan3}. Here, we extend its usage to a more general case, curve extraction from
curve support regions.

We represent the outer boundary of each support region using its Fourier descriptors. Let a complex periodic function,
$u(t)=x(t)+jy(t)=u(t+rT)$, $j=\sqrt{-1}$ for any integer values of $t$ and $r$, represent the outer boundary of the support
region. $T$ is the total number of points in the contour. The complex periodic contour can be approximated by a Fourier series as
\cite{Buck1}:

\begin{equation} \label{eq:fsf2}
\hat{u}(t)=\sum_{n=0}^{T-1}U_n e^{j\frac {2\pi nt} {T}}
\end{equation}

\begin{equation} \label{eq:fsf4}
U_n=\frac {1} {T} \sum_{t=0}^{T-1}u(t) e^{-j\frac {2\pi nt} {T}}
\end{equation}

The curvature of the boundary representation leads to the parametric curve segment extraction from the corresponding support
region. The curvature is a differential geometric entity giving a measure of how rapidly the curve deviates from the tangent line
\cite{Docarmo1}. We find the curvature of the approximated boundary ($\hat{u}(t)=\hat{x}(t)+j\hat{y}(t)$) as:

\begin{equation} \label{eq:curvature}
K(t)= \frac { \frac {d\hat{x}(t)} {dt} \frac {d^2\hat{y}(t)} {dt^2} - \frac {d\hat{y}(t)} {dt} \frac {d^2\hat{x}(t)} {dt^2} } {\left(
(\frac {d\hat{x}(t)} {dt})^2 + (\frac {d\hat{y}(t)} {dt})^2 \right)^{3/2}}
\end{equation}

Given the support region construction and approximation above, the extremal points of this curvature correspond to the endpoints of the
parametric curve segment (on the boundary) to be extracted. There is no algebraic solution to obtain the roots of Eqn.~\ref{eq:curvature}
directly; we solve it numerically.

Assume that we obtain two extremum points $P_1,P_2$, such that each correspond to an endpoint of the curve segment. We divide the
approximated boundary into two parts $x_u+jy_u$ and $x_l+jy_l$ based on these extremum points. These parts have the same parameter such
that $k=[0,P_2-P_1)$ and 0 corresponds to $P_1$, $P_2-P_1$ corresponds to $P_2$ on the boundary. Therefore, they have the same increment
direction. Our final parametric curve segment, $s(k)$ for $k=[0, P_2-P_1)$ is:

\begin{equation} \label{eq:csffinal}
s(k)=\frac{1} {2} \left[\left( x_u(k)+ x_l(k) \right) +j \left( y_u(k)+ y_l(k) \right)\right]
\end{equation}

Depending on the complexity of the scene and the LoG scale parameter selected, some curve support regions may merge. This will
lead to more than two extremum points in the curvature function. For such cases, we first obtain the centroid on the support
region. Then, we obtain all parametric curve segments separately by using each extremum of the curvature and the centroid with
the previous setup as if we have two extremum points.

\section{Experimental Results}

We test our parametric curve segment extraction method on various images. First, we extract parametric curve segments from black
and white images. They give insight on the applicability of our method to the shape information only. Then, we test our method on
grayscale images with diverse characteristics, to emphasize its general applicability.

\subsection{Experiments on Black and White Images}

We start with the rectangle image in Fig.~\ref{fig:rectangle}~(a). We give the parametric curve segments in
Fig.~\ref{fig:rectangle}~(d). As can be seen, our method extracts four parametric curve segments from the rectangle shape as
desired. In this and the following examples, each curve segment is labeled by dots at their endpoints.

We apply the method on two more black and white images. The first image contains an `S' shape. The second image contains
characters `A, B, C, D, E'. We specifically chose these black and white images to observe the performance of our method on
different shape characteristics. Furthermore, to test the robustness of our method, we applied a rotation of $\pi/6$ radians, an
affine transformation as in Eqn.~\ref{eq:affine1}, and an affine transformation followed by a rotation to our images.

\begin{equation} \label{eq:affine1}
\begin{bmatrix}
y'\\
x'\\
\end{bmatrix}
=
\begin{bmatrix}
1&0\\
0.5&1\\
\end{bmatrix}
\times
\begin{bmatrix}
y\\
x\\
\end{bmatrix}
\end{equation}

\begin{figure*}[htbp]
\centering \subfigure{\includegraphics[width=1 in]{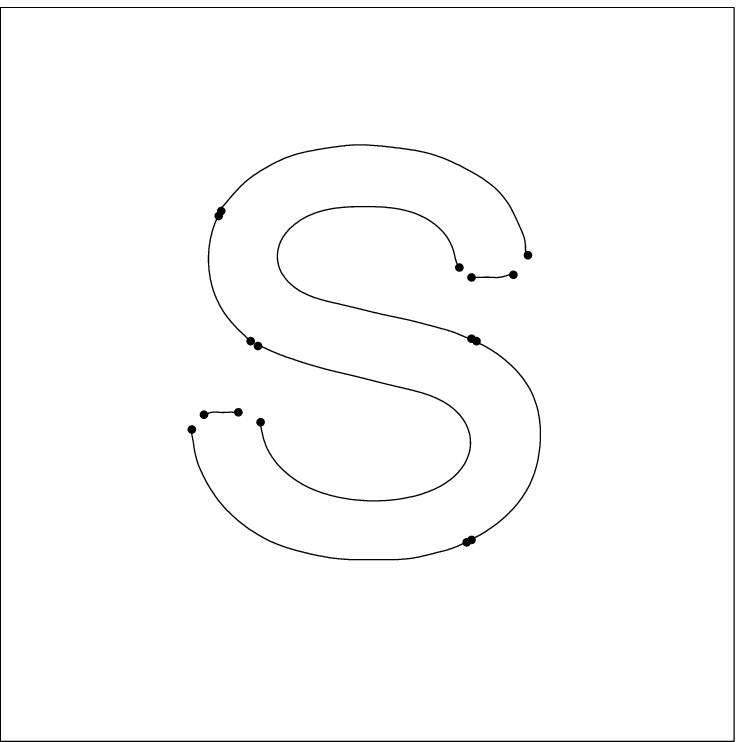}} \subfigure{\includegraphics[width=1
in]{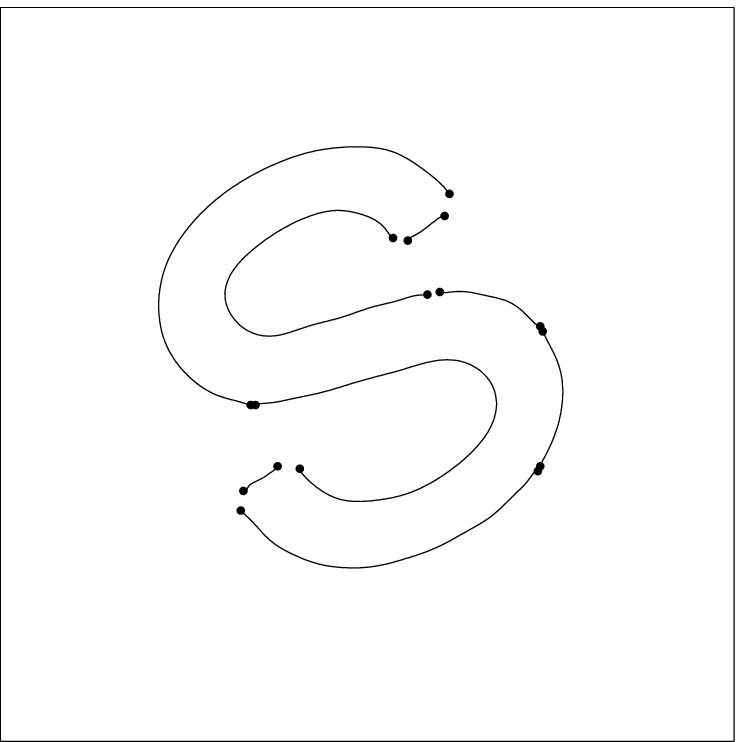}} \subfigure{\includegraphics[width=1 in]{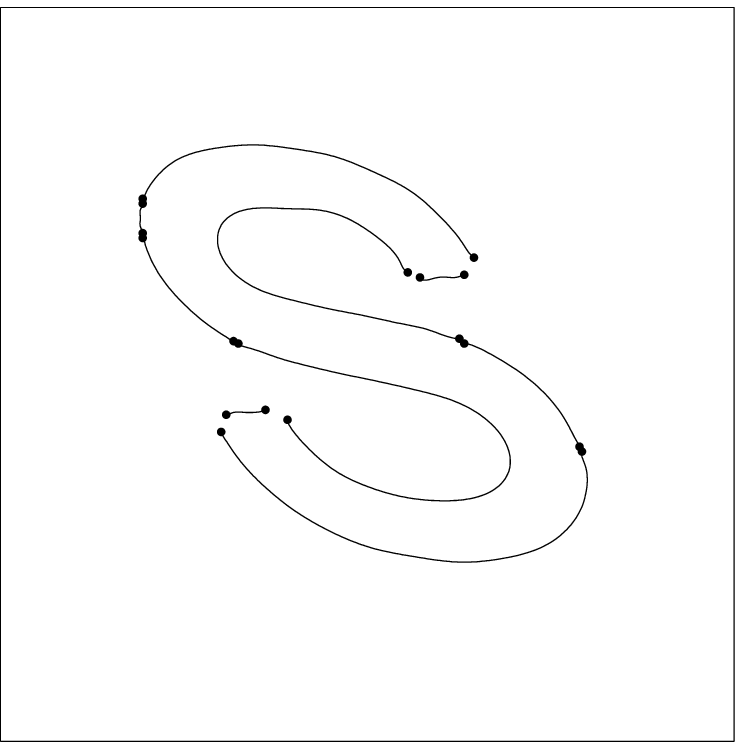}}
\subfigure{\includegraphics[width=1 in]{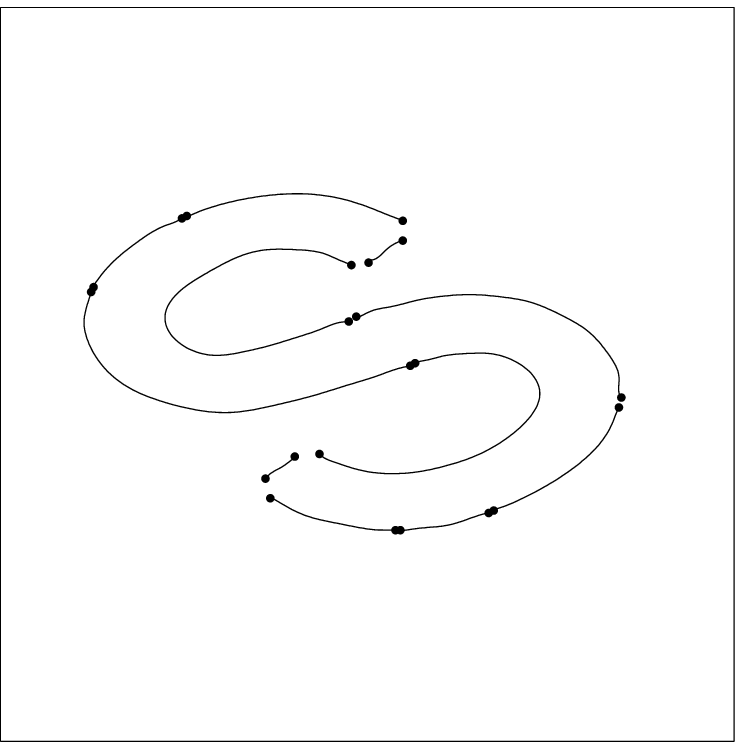}}

\caption{Parametric curve segments obtained from the `S' shape in the original, rotated, affine transformed, affine transformed
and rotated versions (from left to right respectively)} \label{fig:Sshapecurves}
\end{figure*}

For the `S' shape, extracted parametric curve segments are given in Fig.~\ref{fig:Sshapecurves}. The method works fairly well on
the original image. Obtained curve support regions are correct and consistent through rotations and translations. In rotated and
transformed images, there are extra segments, each being a valid curve segment. They can be merged by a postprocessing step.

\begin{figure*}[htbp]
\centering \subfigure{\includegraphics[width=1 in]{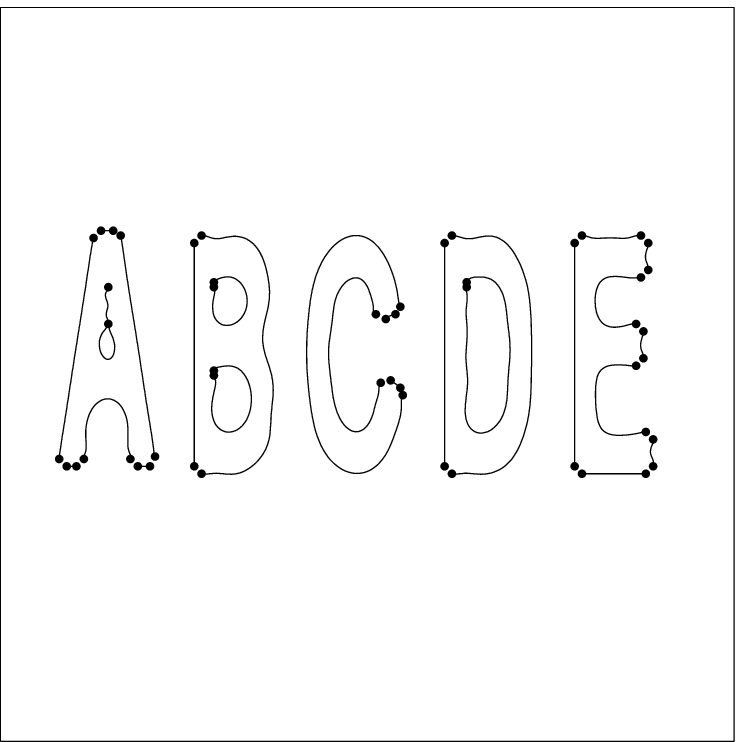}} \subfigure{\includegraphics[width=1
in]{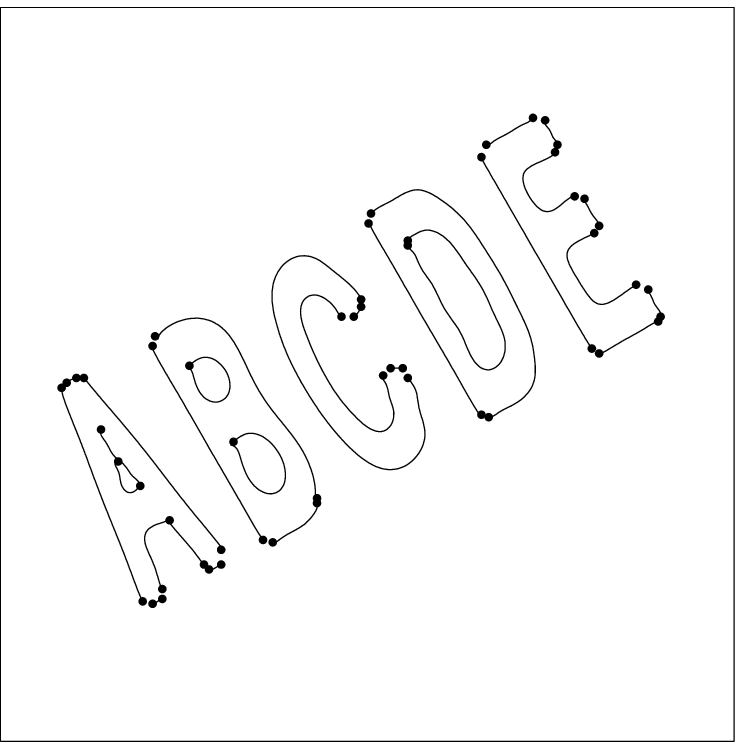}} \subfigure{\includegraphics[width=1 in]{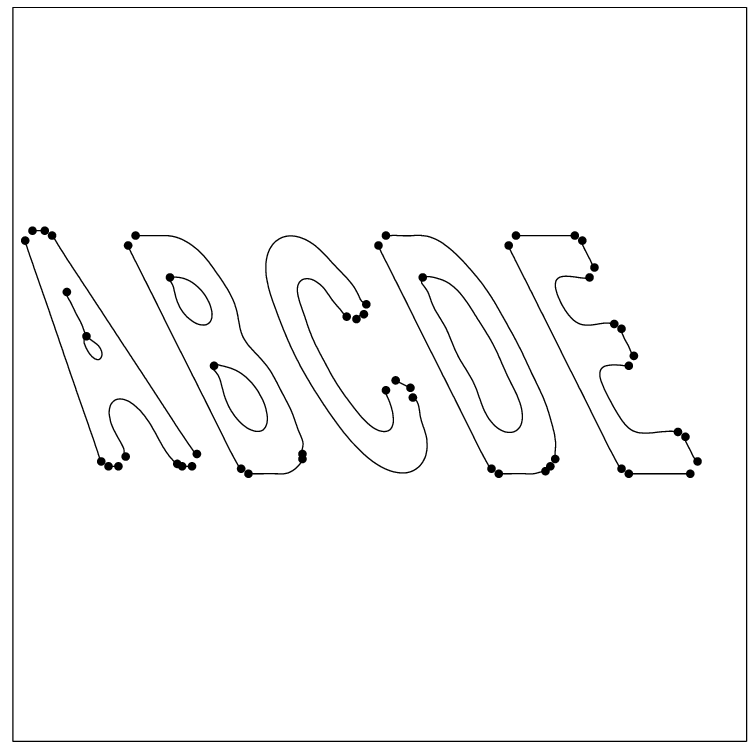}}
\subfigure{\includegraphics[width=1 in]{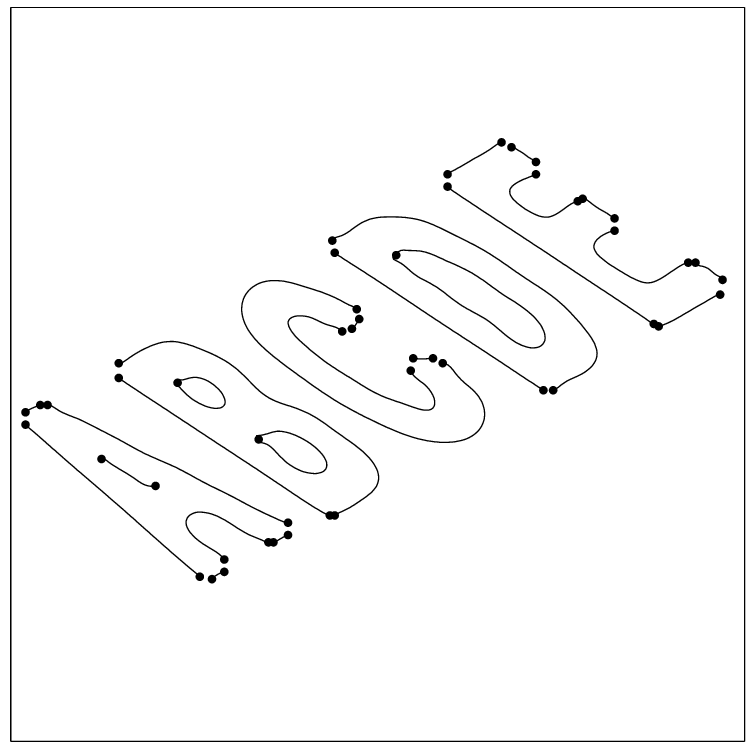}}

\caption{Parametric curve segments obtained from characters in the original, rotated, affine transformed, affine transformed and
rotated versions (from left to right respectively)} \label{fig:characterscurves}
\end{figure*}

As for the characters `A, B, C, D, E', extracted curve segments (with the above variations) are given in
Fig.~\ref{fig:characterscurves}. In this example, `B' has a support region containing both convex and concave parts because of a
large filter scale selection. Since the filter scale is large, such small changes can not be detected in this character.
Similarly, for `E' lines inside the character are combined as one curve. This also occurs from the large filter scale selection,
such that the corners of these lines are merged. Similar to the previous example, extra segmentations can be merged. Besides, all
curve segments are extracted from these characters with success.

We can comment on the black and white image performance of our method as follows. Since we have a rotation invariant filter, ideally there
should be no change on the support regions extracted after rotation. However, we have extra segmentations in the rotated shapes possibly
due to the (discrete) rotation operation. As for the affine transformed and affine transformed and rotated images, we obtain the desired
support regions with some extra segments. As we mentioned previously, these extra segments can be merged easily.

\subsection{Experiments on Grayscale Images}

\begin{figure*}[ht]
\centering

\subfigure{\includegraphics[width=1.55 in]{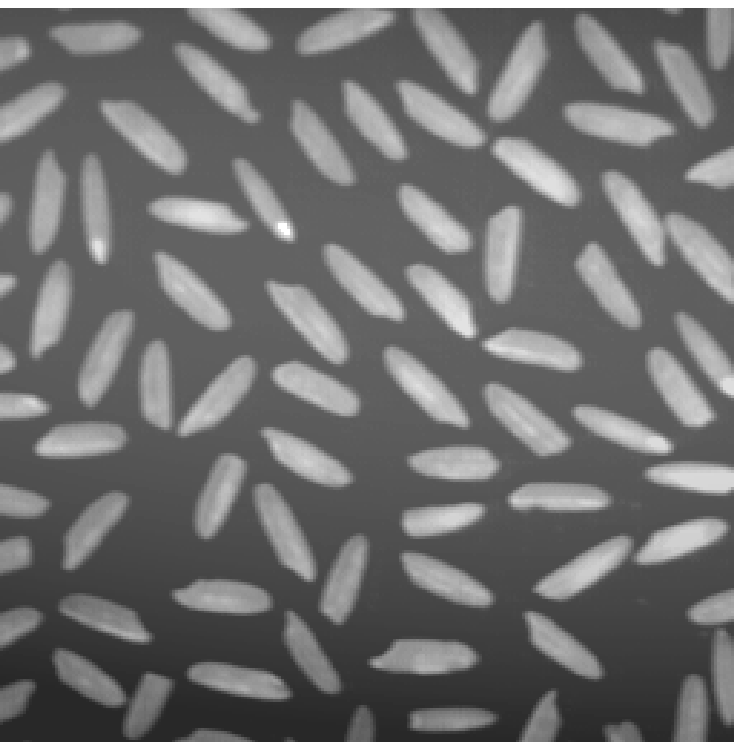}} \subfigure{\includegraphics[width=1.51 in]{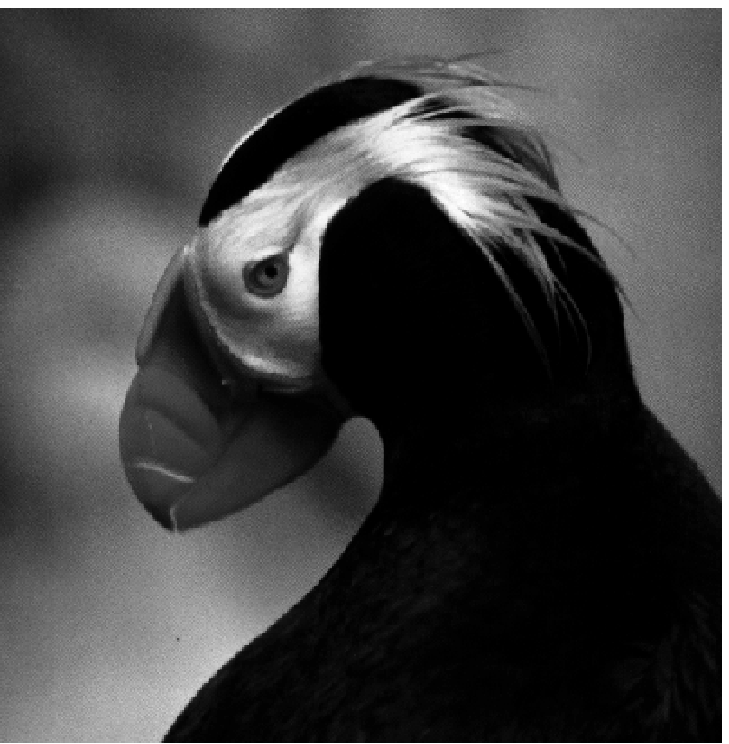}}
\subfigure{\includegraphics[width=1.55 in]{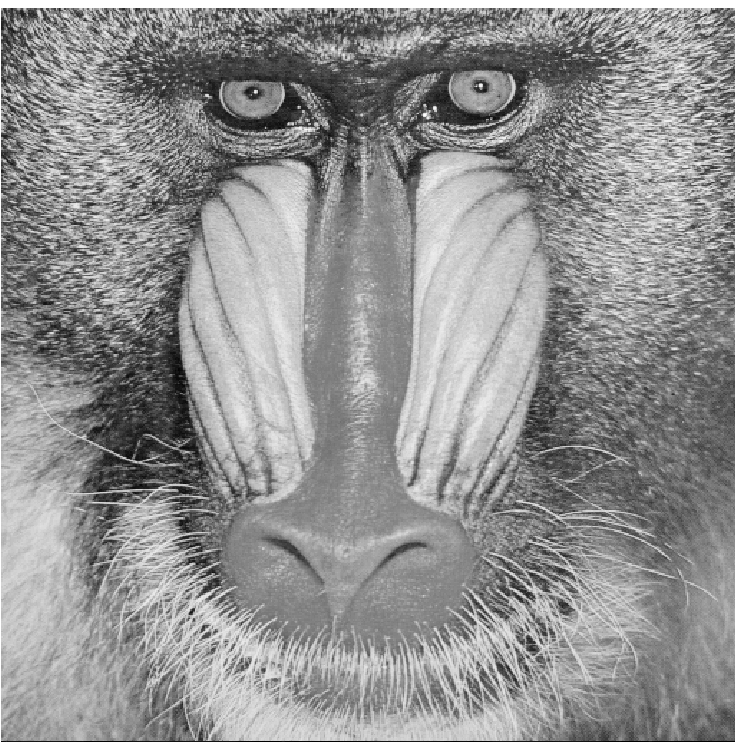}}

\subfigure{\includegraphics[width=1.55 in]{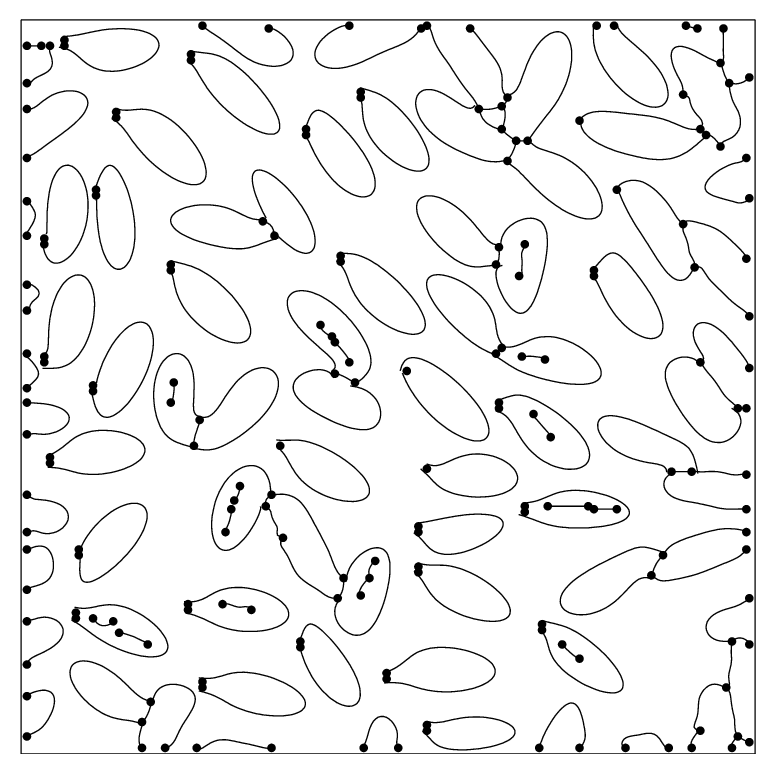}} \subfigure{\includegraphics[width=1.51
in]{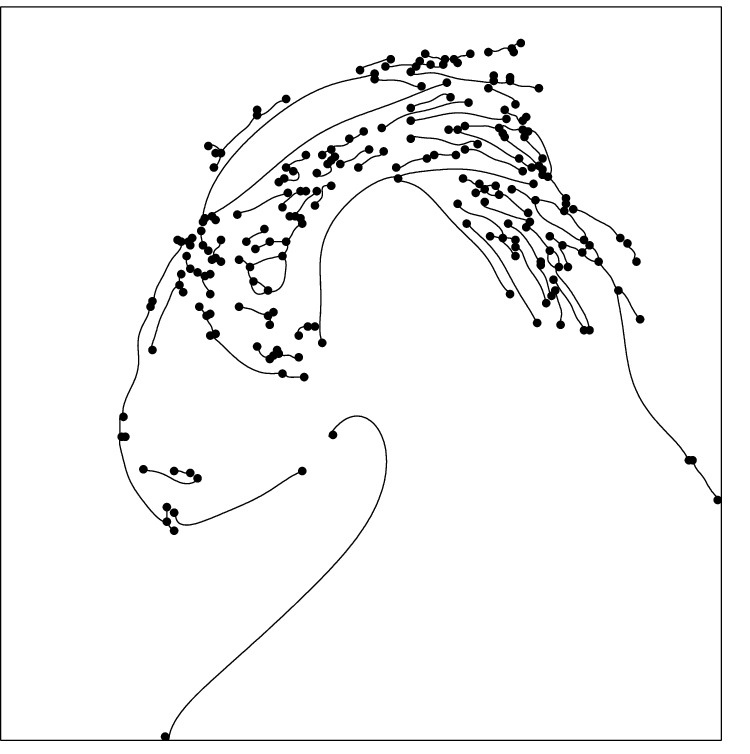}} \subfigure{\includegraphics[width=1.55 in]{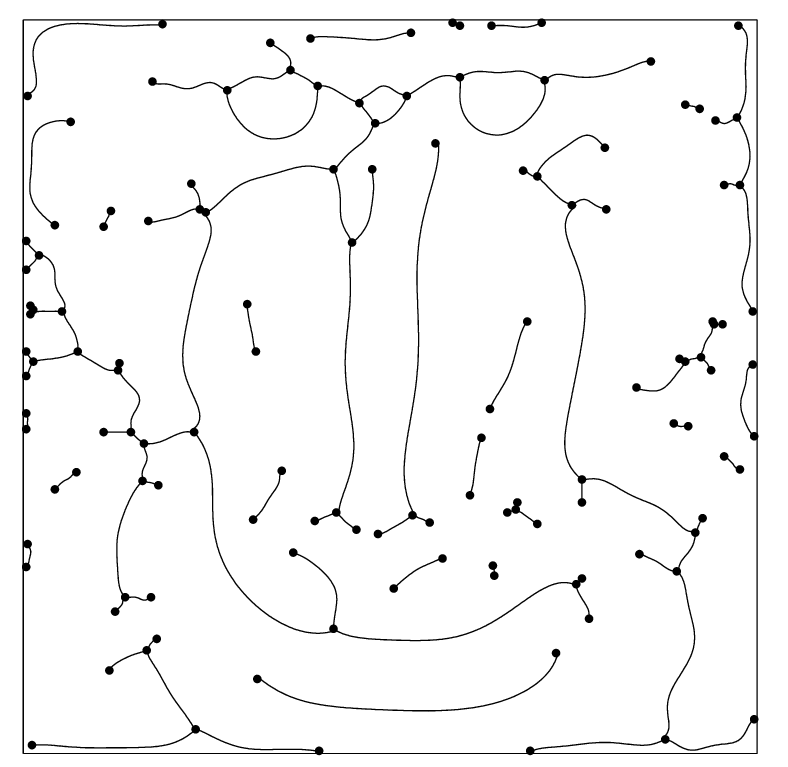}}

\subfigure{\includegraphics[width=1.55 in]{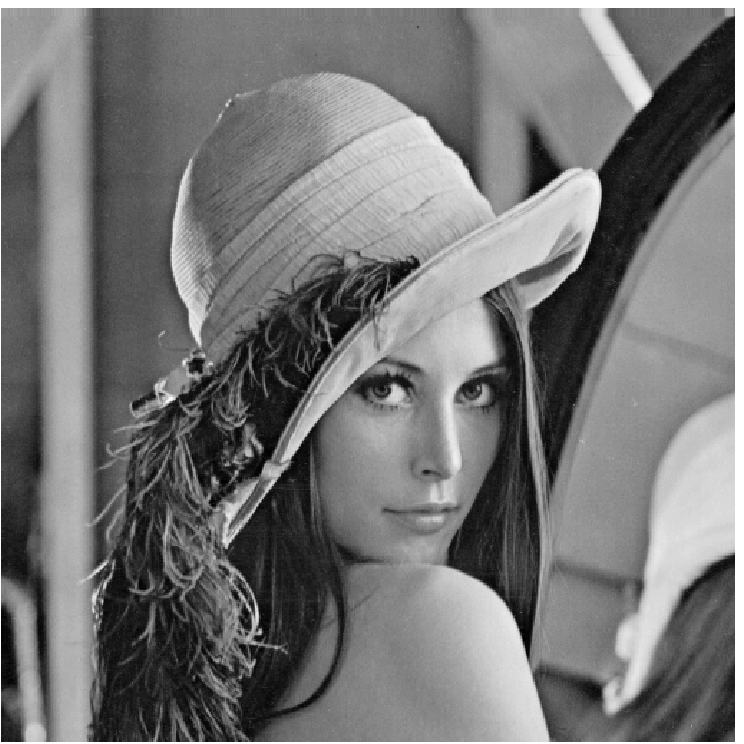}} \subfigure{\includegraphics[width=1.55 in]{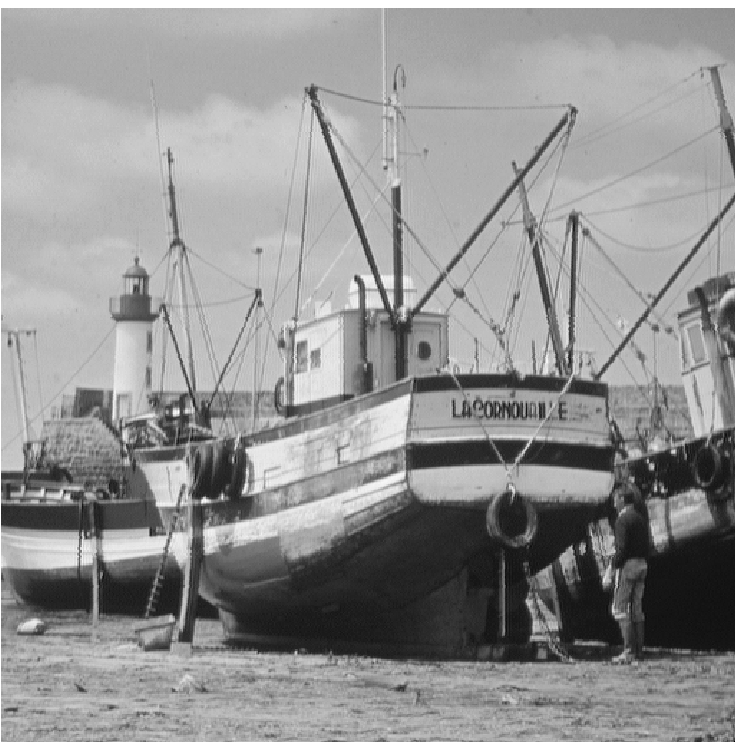}}
\subfigure{\includegraphics[width=1.55 in]{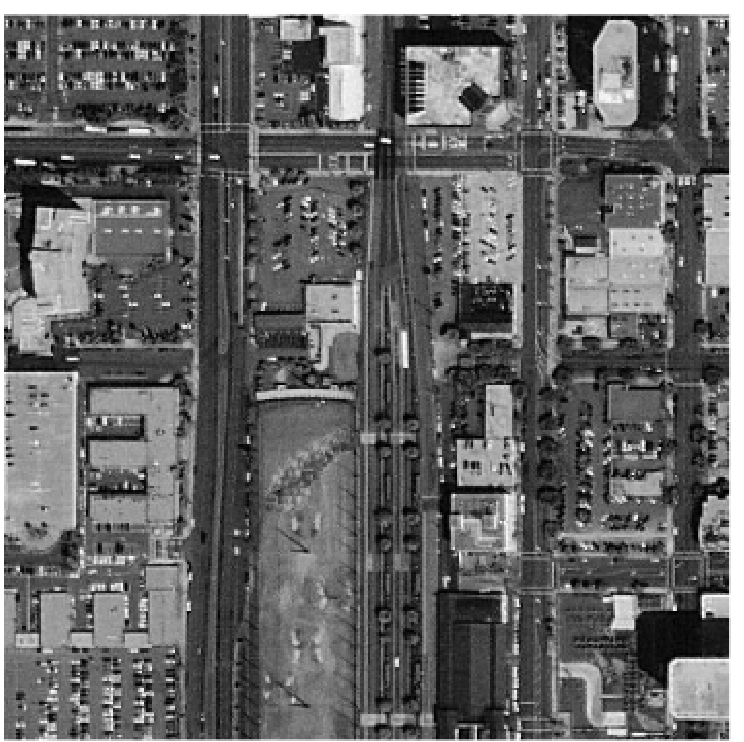}}

\subfigure{\includegraphics[width=1.55 in]{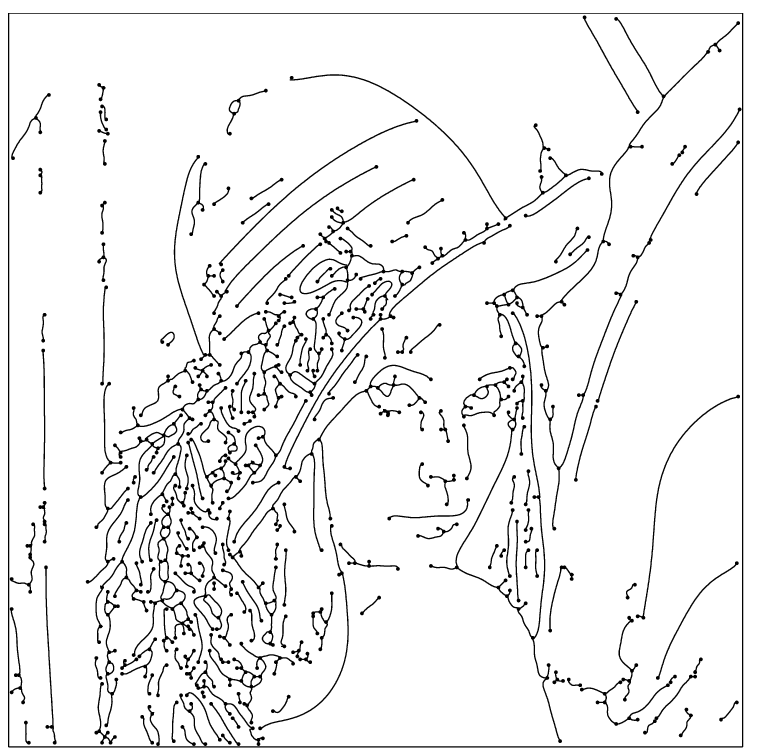}} \subfigure{\includegraphics[width=1.55
in]{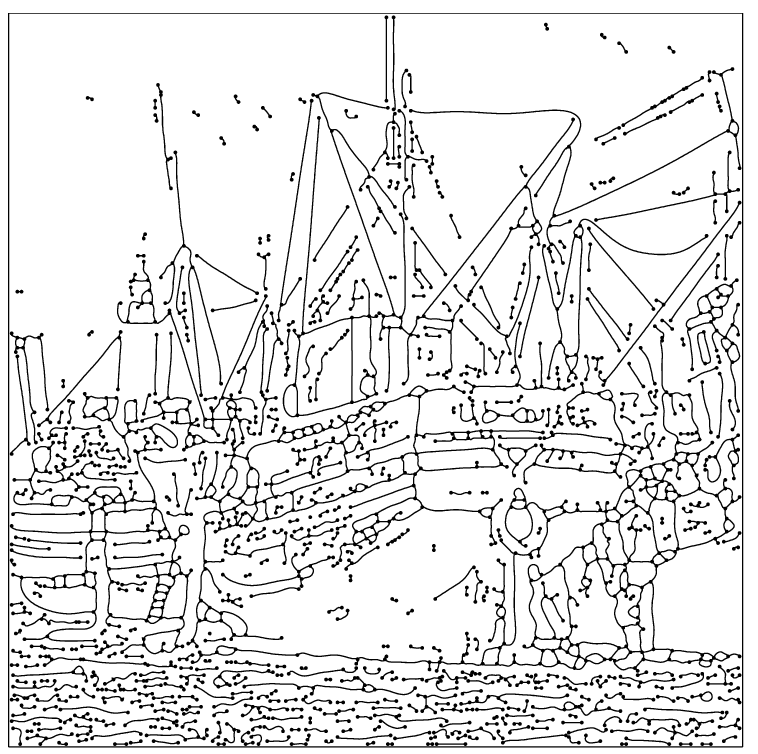}} \subfigure{\includegraphics[width=1.55 in]{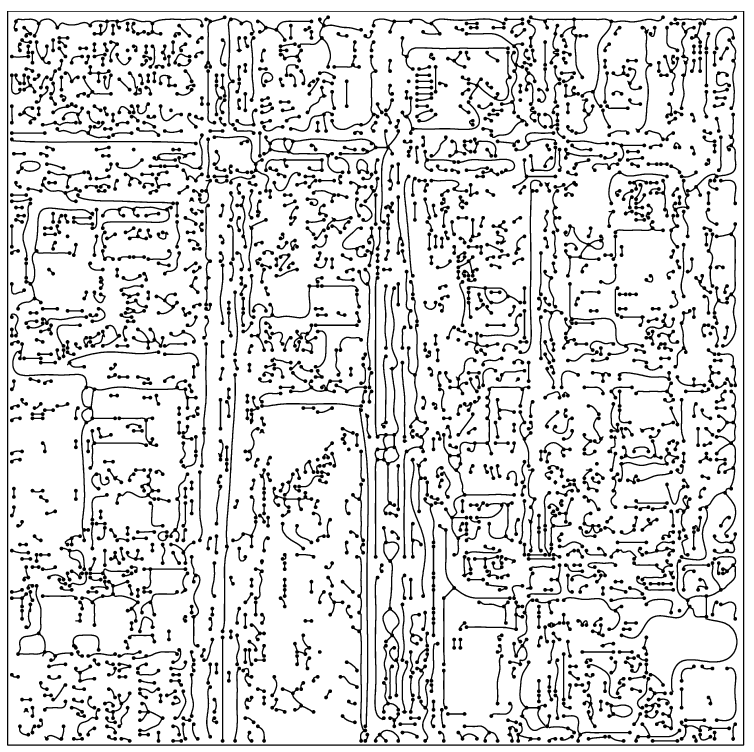}}

\caption{Parametric curve segment extraction from grayscale images.} \label{fig:examples1}
\end{figure*}

Next, we consider the performance of our parametric curve segment extraction method on grayscale images. The first three of these
images (Rice, Puffin, and Mandrill), given in the first row of Fig.~\ref{fig:examples1}, have relatively low content. Our method
locates most of the curve segments (in parametric form) in the low content images with success. In This image set, the Mandrill
image deserves a special consideration. It contains heavy textured regions as well as a general shape. By selecting a large
filter scale, we can detect curve segments (in parametric form) corresponding to the general characteristics of the Mandrill
image among the fine details. Unfortunately, a large scale causes relatively large shift in the location of the detected curve
segments. This also causes some small curve segments to have convex and concave parts. However, the overall performance is
satisfactory.

The last three grayscale images (Lena, Boat, and Urban), given in the third row of Fig.~\ref{fig:examples1}, have high content.
As can be seen, for these images our method works fairly well. Since there are many curve segments in these images, we did not
put heavy dots at the end of each curve segment.

\section{Conclusions}

We introduce a method to extract curve segments (represented in parametric form) directly from the LoG filter response of the
image. Hence, our method combines three basic operations to one by a simple filtering and grouping. Therefore, the proposed
method is potentially faster and simpler than the existing curve extraction, segmentation, and parametric representation methods.
Although the LoG filtering causes a small shift in the location of curves extracted, the general shape of the object is not
affected. Therefore, this method is suitable for most of the recognition and computer vision applications. The experiments
support the usefulness of the proposed method in extracting curve segments both on black and white and grayscale images having
diverse characteristics. For some shapes, the proposed method may give extra segments. However, since we have the parametric
representation of these segments at hand, they can be merged by a postprocessing step.

\bibliographystyle{splncs}

\end{document}